%% file: main.tex
\pdfoutput=1
\documentclass[11pt]{article}
\usepackage[preprint]{acl}
\usepackage{times}
\usepackage{latexsym}
\usepackage[T1]{fontenc}
\usepackage[utf8]{inputenc}
\usepackage{microtype}
\usepackage{inconsolata}
\usepackage{graphicx}
\usepackage{subcaption} 
\usepackage{xcolor}
\usepackage{booktabs}
\usepackage{graphicx}
\usepackage{tcolorbox}
\tcbuselibrary{most}
\usepackage{multirow}
\usepackage{algorithm}
\usepackage{algorithmic}
\usepackage{verbatim}
\usepackage{amsmath}
\usepackage{makecell}
\newtcolorbox{custombox}[1][]{
    colback=gray!10,
    colframe=gray!40,
    fonttitle=\bfseries,
    coltitle=black,
    colbacktitle=gray!60,
    enhanced,
    drop shadow=black!5!white,
    left=3mm,   
    right=3mm,  
    top=3mm,    
    bottom=3mm, 
    toptitle=2mm,    
    bottomtitle=2mm, 
    boxsep=0mm,
    sharp corners=south,
    rounded corners=north,
    title={#1},
    }

\title{Visual Hallucinations of Multi-modal Large Language Models}

\author{Wen Huang\thanks{Equal contributions.} \\
  University of Science \& Technology of China \\
  \texttt{hw2000@mail.ustc.edu.cn} \\\And
  Hongbin Liu\footnotemark[1] \\
  Duke University \\
  \texttt{hongbin.liu@duke.edu} \\
  \AND
  Minxin Guo \\
  The University of Hong Kong \\
  \texttt{xc666@connect.hku.hk} \\\And
  Neil Zhenqiang Gong \\
  Duke University \\
  \texttt{neil.gong@duke.edu} \\}

\newcommand{\ibinospace}{initial VH instance}
\newcommand{\ibis}{initial VH instances~}
\newcommand{\ibisnospace}{initial VH instances}
\newcommand{\ti}{VH instance~}
\newcommand{\tinospace}{VH instance}
\newcommand{\tis}{VH instances~}
\newcommand{\tisnospace}{VH instances}
\newcommand{\hm}{VH mode~}
\newcommand{\hmnospace}{VH mode}
\newcommand{\hms}{VH modes~}
\newcommand{\hmsnospace}{VH modes}
\newcommand{\dallethree}{DALL\textperiodcentered E-3~}
\newcommand{\dallethreenospace}{DALL\textperiodcentered E-3}
\newcommand{\method}{VHTest~}
\newcommand{\methodnospace}{VHTest}
\newcommand{\gpt}{GPT-4V~}
\newcommand{\gptnospace}{GPT-4V}

\newcommand{\gptfournospace}{GPT-4}
\newcommand{\llava}{LLaVA-1.5~}
\newcommand{\llavanospace}{LLaVA-1.5}
\newcommand{\minigpt}{MiniGPT-v2~}
\newcommand{\minigptnospace}{MiniGPT-v2}
\newcommand{\ttim}{text-to-image generative model~}
\newcommand{\ttimnospace}{text-to-image generative model}
\newcommand{\ttimnospaceupper}{Text-to-image Generative Model}

\begin{document}
\maketitle

\input{0_Abstract} 
\input{1_Introduction}
\input{2_Problem_Definition}
\input{3_Method}
\input{4_Experiment}
\input{5_Related_work}

\input{6_Conclusion}
\bibliography{ref}
\appendix
\input{X_suppl}

\end{document}

%% file: 0_Abstract.tex
\begin{abstract}
\emph{Visual hallucination (VH)} means that a multi-modal LLM (MLLM) imagines incorrect details about an image in visual question answering. Existing studies  
find VH instances only in existing image datasets, which results in biased understanding of MLLMs' performance under VH due to limited diversity of such VH instances. In this work, we propose a tool called \method to generate a diverse set of VH instances. Specifically, \method finds some initial VH instances in existing image datasets (e.g., COCO), generates a text description for each VH mode, and uses a \ttim (e.g., \dallethreenospace) to generate VH images based on the text descriptions. We collect a benchmark dataset with 1,200 VH instances in 8 VH modes using \methodnospace. We find that existing MLLMs such as \gptnospace, \llavanospace, and \minigpt hallucinate for a large fraction of the instances in our benchmark. Moreover, we find that fine-tuning an MLLM using our benchmark dataset reduces its likelihood to hallucinate without sacrificing its performance on other benchmarks.  Our benchmarks are publicly available: 
\href{URL}{https://github.com/wenhuang2000/VHTest}. 
\end{abstract}

%% file: 1_Introduction.tex
\section{Introduction}
\label{sec:intro}
A multi-modal LLM  (MLLM)~\cite{yang2023dawn,zhu2023minigpt,chen2022visualgpt,huang2023language,tiong2022plug}  generates a \emph{text response} for a given \emph{image} and \emph{question}. An MLLM typically comprises three components: a vision encoder, a vision-language connector, and an  LLM. The vision encoder (e.g., CLIP~\cite{radford2021learning}) converts an image into an embedding vector. The vision-language connector projects an image embedding vector into the LLM's word embedding space. The projected vector is concatenated with the token embeddings of the question to form an input to the LLM, which generates a text response.

\begin{figure}[!t]
    \centering
    \includegraphics[width=0.48\textwidth]{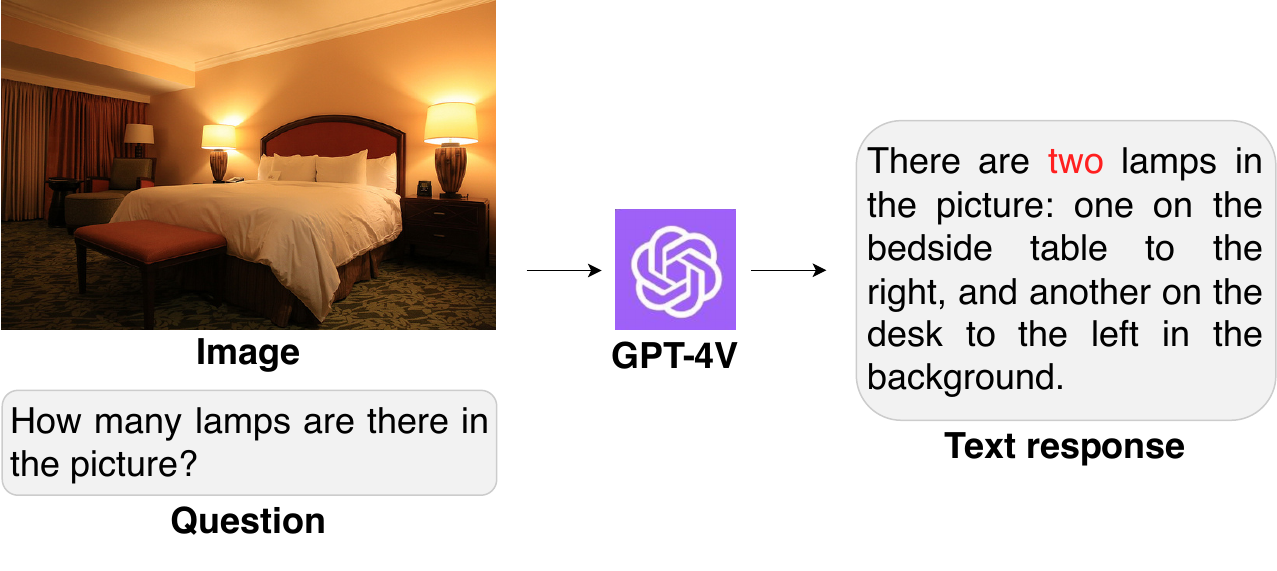}
    \caption{An example of MLLM's visual hallucination. The text in {\color{red}{RED}} highlights the hallucinated detail in the image, where there are three lamps.} 
    \vspace{-5mm}
    \label{fig:vh_example}
\end{figure}

However, MLLMs often generate text responses containing factually incorrect details about an image, known as \emph{visual hallucination (VH)}~\cite{li2023evaluating,liu2024survey}. Figure~\ref{fig:vh_example} shows an example where the MLLM hallucinates two lamps, contradicting the three lamps in the image. VHs in MLLMs pose obstacles to developing safe and trustworthy AI, which is emphasized in a recent U.S. Executive Order calling for rigorous testing to address potential harms from advanced AI systems~\cite{Biden2023AI}.

Prior works have tried to benchmark MLLMs' VHs related to object existence~\cite{li2023evaluating,liu2024mitigating}, optical character recognition (OCR),  object counting,  object positions comparing~\cite{fu2023mme}, orientation, and viewpoint~\cite{tong2024eyes} (concurrent to ours). However, they collect VH images only from existing image datasets like COCO~\cite{lin2014microsoft}. This limits the diversity of VH images since they can only find a limited number of them. Moreover, existing image datasets may have been used to pre-train an MLLM, leading to data contamination~\cite{jacovi2023stop,sainz2023nlp}. As a result, such VH images lead to a biased understanding of an MLLM's performance, e.g., an MLLM is incorrectly concluded to perform well under VH.

\begin{figure*}[!t]
  \centering
  \includegraphics[width=1.0\textwidth]{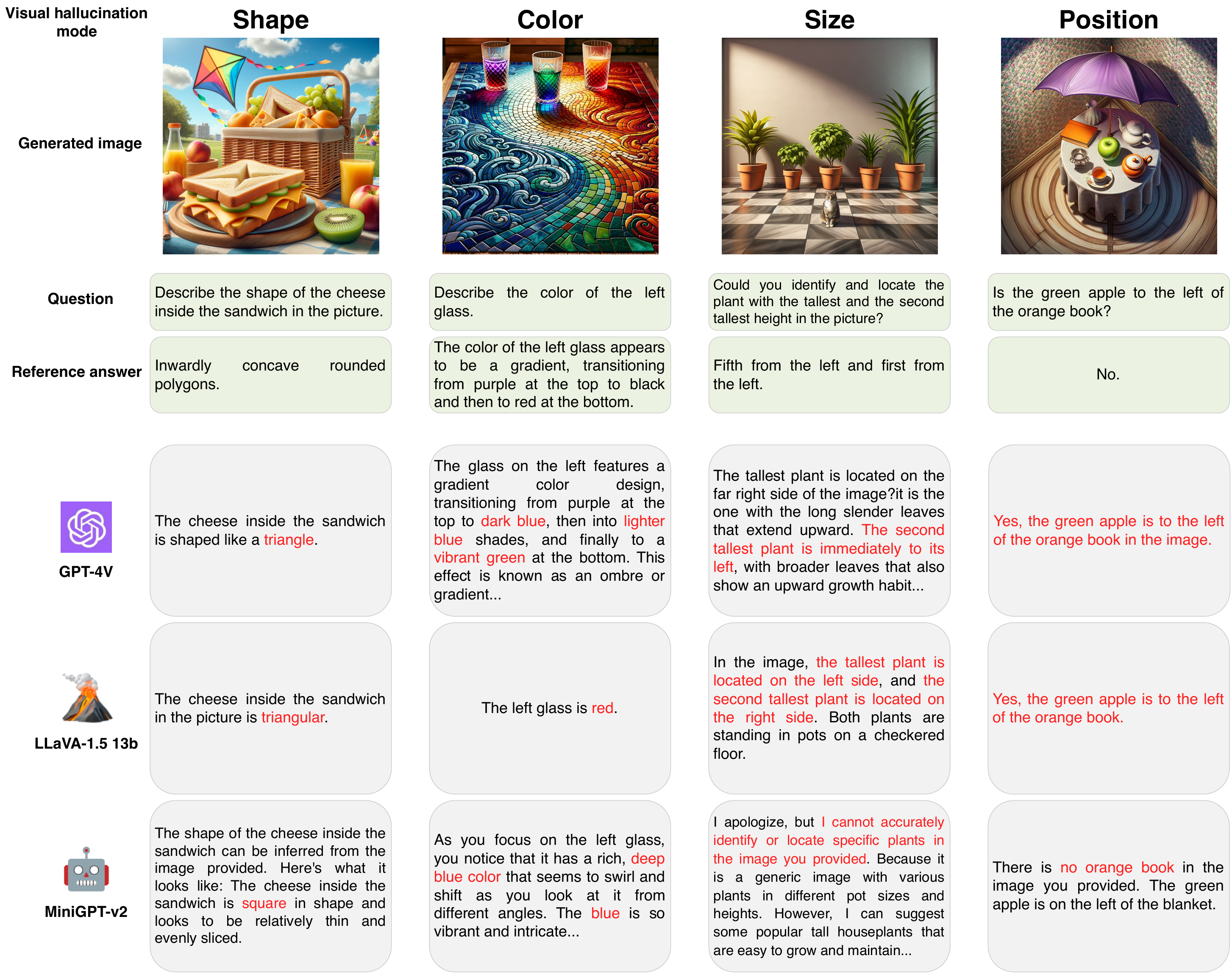}
  \caption{Example VH instances generated by our \method and the text responses of three MLLMs for them. Figure~\ref{fig:other_four_examples_for_the_other_VH_modes} in Appendix shows examples for the other four VH modes.} 
  \label{fig:successful_visual_instances}
\end{figure*}

\paragraph{Our Work}  
We propose \methodnospace, a tool that generates VH instances through adversarial construction to expose visual hallucinations within MLLMs. A \ti is a triple (an image, a question, a reference answer). Our \method has three key steps. Step \uppercase\expandafter{\romannumeral1} finds initial VH instances using existing image datasets like COCO. Specifically, we first identify image pairs with high CLIP embedding similarity but low DINO v2~\cite{oquab2023dinov2} embedding similarity. Such image pairs have contradictory similarities from two powerful vision encoders, indicating potential VHs. We note that Step \uppercase\expandafter{\romannumeral1} is also used in a concurrent work MMVP~\cite{tong2024eyes}. Unlike MMVP, which directly collects initial VH instances to build their benchmarks, we treat them as the raw materials for further adversarial construction. We then manually design questions and reference answers for these images to obtain initial VH instances. Step \uppercase\expandafter{\romannumeral2} generates a text description for a VH mode derived from the initial VH instances. A text description describes visual properties of VH images that are likely to cause MLLMs to hallucinate. Finally, Step \uppercase\expandafter{\romannumeral3} uses a \ttim (e.g., DALL-E 3) to generate new images based on the text descriptions. Moreover, based on some templates, we design questions and reference answers for the generated images to construct VH instances.

Using our~\methodnospace, we construct a new benchmark dataset for evaluating VHs in MLLMs. Our benchmark contains 1,200 VH instances covering 8 VH modes. The 8 VH modes are related to \emph{existence}, \emph{shape}, \emph{color}, \emph{orientation}, \emph{OCR}, \emph{size}, \emph{position}, and \emph{counting} of visual objects in an image. Note that shape and size VH modes are formulated by us, while the other 6 VH modes were also considered in prior~\cite{yang2023dawn} and concurrent~\cite{tong2024eyes} studies. Figure \ref{fig:successful_visual_instances} and Figure \ref{fig:other_four_examples_for_the_other_VH_modes} show some VH instances generated by~\methodnospace.

We comprehensively evaluate state-of-the-art MLLMs, including \gptnospace, \llavanospace, and \minigpt on our benchmark. Our results show that MLLMs hallucinate for a large fraction of the VH instances in our benchmark. For example, \gptnospace, \llavanospace, and \minigpt only achieve overall accuracy of 0.383, 0.229, and 0.075 on our benchmark, respectively. We also find that   MLLMs have different performance across \hmsnospace. For example, \gpt is most prone to orientation VH with 0.153 accuracy; while \llava and \minigpt are most susceptible to OCR VH with 0.127 and 0.000 accuracy, respectively.

Finally, we show that fine-tuning an MLLM using our benchmark dataset mitigates VH. Specifically, we divide our benchmark into the training/testing splits with a ratio 80\%/20\%. We then fine-tune the \llava model on the training split. After fine-tuning, we evaluate the model on the testing split. Our results show fine-tuning reduces the likelihood for an MLLM to hallucinate. For example, in position \hmnospace, the fine-tuned \llava gains 0.200 accuracy from 0.333 to 0.533. Moreover, fine-tuning maintains model performance on other benchmark datasets.

%% file: 2_Problem_Definition.tex
\section{Definitions}
\label{sec:definition}

\paragraph{VH Modes} We can categorize visual properties of objects in an image into \emph{individual} properties, which can be attributed to individual objects (e.g., existence, shape, color, orientation, and OCR), and \emph{group} properties, which emerge from comparisons across multiple objects (e.g., relative size, relative position, and counting). Based on such categorization, we have 8 VH modes as follows. In particular, each~\hm occurs when an MLLM's text response is factually incorrect with respect to the corresponding visual property in an image.

\begin{enumerate}

    \item \textbf{Existence VH}: 
     \(O\) is the set of objects in image \(I\), and \(O'\) is the set of objects an MLLM identifies in \(I\). Existence VH occurs if: $\exists o_i \in O \, \, \:\:s.t.\:\: o_i \notin O' \:\:\:or\:\:\: \exists o'_j \in O' \:\:s.t.\:\: o'_j \notin O$. In other words, the MLLM misses at least one object in \(I\) or fabricates at least one nonexistent object.

    \item \textbf{Shape VH}:
    Let \(S=\{s(o_i)\}_{i=1}^{n}\) denote the list of shapes for objects \(\{o_i\}_{i=1}^{n}\) in $I$, and \(S'=\{s'(o_i)\}_{i=1}^{n}\) is the corresponding list of shapes identified by an MLLM. A shape VH occurs if: $\exists o_i \:\:s.t.\:\: s(o_i) \neq s'(o_i)$. Intuitively, the MLLM fails to accurately describe the shape of at least one object in \(I\).
    
    \item \textbf{Color VH}: 
    Let \(C=\{c(o_i)\}_{i=1}^{n}\) denote the list of colors for objects \(\{o_i\}_{i=1}^{n}\) in $I$, and \(C'=\{c'(o_i)\}_{i=1}^{n}\) is the corresponding list of colors identified by an MLLM. A color VH occurs if the MLLM fails to accurately identify the color of at least one object in \(I\). Formally, we have: $\exists o_i \:\:s.t.\:\: c(o_i) \neq c'(o_i)$. 
    
    \item \textbf{Orientation VH}: 
    An LLM fails to precisely recognize the facing orientation of at least one object in an image.

    \item \textbf{OCR VH}: An MLLM fails to accurately identify at least one character in an image.
    
    \item \textbf{Size VH}: An MLLM fails to accurately compare the relative sizes of multiple objects in an image.
    
    \item \textbf{Position VH}: An MLLM fails to accurately identify spatial relationships between objects in an image. 
    
    \item \textbf{Counting VH}: An MLLM exhibits a counting VH mode when it cannot accurately enumerate the number of objects in an image. 

\end{enumerate}

\paragraph{VH Instance} An VH instance is a triple $\{x_i, x_t, y_r\}$, where $x_i$ is an image, $x_t$ is a question, and $y_r$ is a reference answer. We say a VH instance \emph{succeeds} for an MLLM if and only if the MLLM's text response for $x_i$ and $x_t$ is factually incorrect compared to the reference answer $y_r$.
 For instance, in the example shown in Figure~\ref{fig:vh_example}, the reference answer is ``three lamps'', while the MLLM's text response indicates two lamps. 

%% file: 3_Method.tex
\section{Our \method}

\begin{figure*}[!t]
    \centering
    \includegraphics[width=1\textwidth]{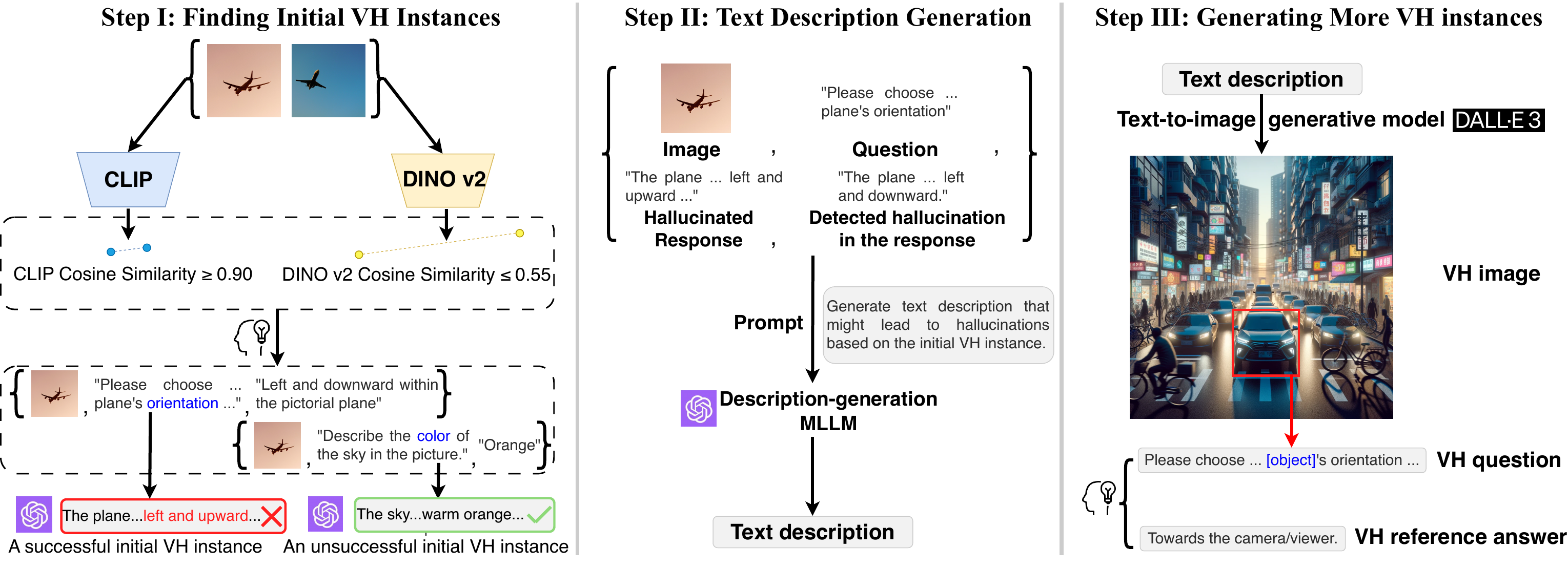}
    \caption{Pipeline of our \methodnospace. The human-head symbol means a human worker manually generates a question-answer pair for an image.}
    \label{fig:figure2_pipeline}
    \vspace{-2mm}
\end{figure*}

\subsection{Step \uppercase\expandafter{\romannumeral1}: Finding Initial VH Instances}
\label{subsec:Step_I}

Since the CLIP vision encoder is the backbone of many popular MLLMs such as LLaVA~\cite{liu2023improved}, LLaMA-Adapter~\cite{gao2023llamaadapter}, and mPLUG-Owl~\cite{ye2023mplugowl2}, we leverage CLIP to find initial VH instances. Our goal is to find images in an existing image dataset (e.g., COCO)  that are incorrectly embedded by the CLIP vision encoder to have high similarity,  despite differences in visual semantics. Such images may lead to VH for MLLMs due to their incorrect embeddings.

Specifically, given an image pair $(x_1, x_2)$, we compute the cosine similarity between their CLIP embedding vectors, i.e., $cos(f_C(x_1), f_C(x_2))$, where $f_C(\cdot)$ is an embedding vector output by CLIP. Moreover, we also use  DINO v2~\cite{oquab2023dinov2} as a reference vision encoder to compute their embedding vectors and compute their cosine similarity, i.e., $cos(f_D(x_1), f_D(x_2))$, where $f_D(\cdot)$ is an embedding vector output by DINO v2. Then, we find image pairs that have large cosine similarity under CLIP but small cosine similarity under DINO v2. In particular, we find image pairs that satisfy $cos(f_C(x_1), f_C(x_2)) \geq 0.9$ and $cos(f_D(x_1), f_D(x_2)) \leq 0.55$. Such image pairs are our candidates.

Among the candidates, we further select the top 200 pairs with the largest cosine similarity under CLIP. Moreover, we manually design questions and reference answers to form 800 initial VH instances, where 100 for each VH mode and an image may be used in multiple VH instances. Finally, we test an MLLM (called \emph{testing MLLM}, e.g., \gptnospace) on them. With manual verification,  \gpt hallucinates on 204 of them (called successful initial VH instances). Table~\ref{tab:initial_coco}  shows the number of successful initial instances in each VH mode.

\subsection{Step\,\uppercase\expandafter{\romannumeral2}: Text Description Generation}
\label{subsec:Step_II}

Given the initial VH instances, we use an MLLM (called \emph{description-generation MLLM}, e.g., \gptnospace) to generate a text description for each VH mode. The text description aims to guide a \ttim (in Step \uppercase\expandafter{\romannumeral3}) to generate more images that are likely to trigger VH in MLLMs.

\paragraph{Using a Successful Initial VH Instance}  We first describe how to leverage prompt engineering to generate a text description based on one successful initial VH instance. Specifically, we construct an example comprised of: 1) the VH instance's image and question, 2) the testing MLLM's hallucinated response, and 3) the detected hallucination in the response. We then add an additional prompt asking the description-generation MLLM to generate a text description that explains potential causes underlying the observed VH and describes how to generate more images. We show a summary version of our prompt in Figure~\ref{prompt:w_initial_VH_instance} while the full prompt is shown in Figure~\ref{prompt:w_initial_VH_instance_full} in Appendix.   

\begin{figure}[!h]
\small
\centering
\begin{custombox}[Summary version]
PROVIDE: \{image, question, hallucinated response, detected hallucination in the response\} \\
PROMPT: \  [Through the example, explain the potential causes of such hallucination and how to produce more images and questions.]
\end{custombox}
\caption{Summary of our prompt to generate a text description based on a successful initial VH instance.}
\label{prompt:w_initial_VH_instance}
\end{figure}

\paragraph{Using an Unsuccessful Initial VH Instance} We also design a prompt to generate a text description based on an unsuccessful initial VH instance for which the testing MLLM does not hallucinate. Such prompt is needed when not enough successful initial VH instances are available for a VH mode. Our idea is to construct a prompt with some \emph{hypothetical hallucinated detail} regarding the  VH image. Specifically, we construct an example comprised of: 1) the VH instance's image and question, 2) a hypothetical hallucinated response, and 3) the detected hallucination in the hypothetical response. Given the example, we also add an additional prompt like we discussed above.  We show a summary version of our prompt in Figure~\ref{prompt:w/o_initial_VH_instance} while the full prompt is shown in Figure~\ref{prompt:w/o_initial_VH_instance_full} in Appendix.   

\begin{figure}[!h]
\small
\centering
\begin{custombox}[Summary version]  
PROVIDE: \{image, question, hypothetical hallucinated response, detected hallucination in the hypothetical response\}\\
PROMPT: \  [Through the example, explain the potential causes of the hallucination and how to generate more images and questions.]
\end{custombox}
\caption{Summary of our prompt to generate a text description based on an unsuccessful initial VH instance.}
\label{prompt:w/o_initial_VH_instance}
\end{figure}

\paragraph{Integrating Multiple VH Instances} 
To increase diversity of our text description, for each VH mode, we generate 10 text descriptions based on  10 successful initial VH instances. If a VH mode has less than 10 successful initial VH instances, we generate the remaining text descriptions based on unsuccessful ones. Then, for each VH mode, we use a prompt in Figure~\ref{prompt:text_description_intergration} in Appendix for a description-generation MLLM to summarize the 10 text descriptions as the final one. Appendix~\ref{textdescriptiongenerated} shows our generated text description for each VH mode.

\subsection{\hspace{-2.8mm}Step\,\uppercase\expandafter{\romannumeral3}:\,Generating\,More\,VH\,Instances}
\label{subsec:Step_III}

We generate more \tis based on the text descriptions. Recall that a \ti consists of an image, a question, and a reference answer. Therefore, we describe how to generate each component in the following.

\paragraph{VH Image} We use a \ttim (e.g., \dallethree in our experiments) to generate VH images based on the text descriptions. Specifically, to generate an image in a VH mode, 
we append the corresponding text description to the prompt in Figure~\ref{prompt:VH_image_generation} in Appendix for the \ttim to generate an image.

\paragraph{VH Question}
Given a VH image, we prepare a question based on it. Specifically, we leverage object-driven templates to create diverse and relevant VH questions. For instance, ``Describe the shape of the [object] in the picture.'' is a template for the shape VH mode. We curate question templates for each VH mode and they are shown in Appendix~\ref{sec:question_templates}.  Given a template for a VH mode, a human worker generates a question via manually analyzing the objects in the VH image. For instance, the human worker may replace the [object] as ``pear'' in the template above when the VH image contains a pear. The human worker verifies that the question should have a non-ambiguous answer based on the VH image. If no questions with non-ambiguous answers can be constructed, the VH image is discarded.

\paragraph{VH Reference Answer}
Given a VH image and a question, a human worker also provides a factually correct answer as a reference answer via manually analyzing the VH image. The triple (image, question, reference answer) is a VH instance.

\subsection{Benchmark Construction}
We use~\method to build two benchmarks. Specifically, we find initial VH instances in COCO~\cite{lin2014microsoft} and we generate 150 \tis for each \hmnospace, which results in 1,200 \tis across 8 \hms in total. This benchmark consists of  ``open-ended question'' (OEQ), which requires manually labeling the responses of MLLMs when testing them on the benchmark. 
Therefore, to facilitate automatic evaluation when testing MLLMs, we also construct a closed-ended  ``yes/no question'' (YNQ) version of the benchmark.

Specifically, for each \tinospace, we convert the open-ended question into a binary ``yes/no'' question. For instance, the open-ended question ``Describe the shape of the pear in the picture.'' is converted into ``Is the shape of the pear in the picture a square?''. Moreover, the reference answer is converted into yes or no. We construct the binary questions to ensure that the YNQ benchmark has a 50/50 split of ``yes''/``no'' reference answers. Moreover, to ensure quality of our benchmarks, we measure the inter-annotator agreement to evaluate the consistency among different human annotators, provided in Appendix~\ref{subsec:inter_annotator_agreement}.

Our benchmark construction took approximately 300 human-hours in total.

%% file: 4_Experiment.tex
\section{Experiments}
\label{sec:experiments}

\begin{table*}[!t]
\centering
\small
\caption{Accuracy of \gpt on the initial VH instances from COCO. Each VH mode has 100 initial VH instances.}
\label{tab:initial_coco}
\begin{tabular}{@{}cccccccccc@{}}
\toprule
& Existence & Shape & Color & Orientation & OCR   & Size  & Position & Counting & Average \\ \midrule
Accuracy & 0.880     & {1.000} & 0.920 & 0.280       & 0.600 & {0.980} & 0.700    & 0.600    & 0.745   \\  
\makecell{Number of successful\\ \ibisnospace} & 12        & 0     & 8     & 72          & 40                    & 2     & 30       & 40       & 204 \\ \bottomrule
\end{tabular}
\end{table*}

\begin{table}[!t]
\small
\centering
\caption{Accuracy of \gptnospace, \llava and \minigpt on  our OEQ benchmark. }
\label{tab:main_acc_performance}
\resizebox{\linewidth}{!}{ 
\begin{tabular}{@{}ccccc@{}}
\toprule
& \gptnospace & \llavanospace & \minigptnospace & Average \\ \midrule
Existence             & 0.427                      & 0.240                        & 0.013                          & 0.227   \\ 
Shape                 & 0.487                      & 0.167                        & 0.093                          & 0.249   \\ 
Color                 & 0.460                      & 0.267                        & 0.053                          & 0.260   \\ 
Orientation           & {0.153}                      & 0.140                        & 0.127                          & {0.140}   \\ 
OCR & 0.367                      & {0.127}                        & {0.000}                          & 0.164   \\ 
Size                  & 0.413                      & 0.353                        & 0.140                          & 0.302   \\ 
Position              & 0.547                      & 0.347                        & 0.147                          & 0.347   \\ 
Counting              & 0.213                      & 0.193                        & 0.027                          & 0.144   \\ \midrule
Average               & 0.383                      & 0.229                        & 0.075                          & 0.229   \\ \bottomrule
\end{tabular}
}
\end{table}

\subsection{Experimental Setup}

\paragraph{MLLMs} We evaluate three state-of-the-art MLLMs on our benchmarks: \gpt with its ``gpt-4-vision-preview'' version, \llavanospace-13b~\cite{liu2023improved}, and \minigpt~\cite{chen2023minigptv2}.

\paragraph{Evaluation Metric} We use \emph{accuracy} as an evaluation metric. Given an MLLM, we use it to produce a text response for each VH instance (more precisely, the image and question in a VH instance) in our OEQ benchmark; we manually analyze the text responses and compare with the reference answers; and accuracy is the fraction of the VH instances for which the MLLM's text responses are factually correct. For the YNQ benchmark, we can automatically calculate the accuracy of an MLLM since the reference answers are just yes or no. Note that a smaller accuracy indicates that an MLLM is more likely to hallucinate on our benchmarks, which shows that our \method is better at generating successful VH instances.

\subsection{Testing VH in MLLMs}

\paragraph{Initial VH Instances are Insufficient} Table~\ref{tab:initial_coco} shows the accuracy of \gpt on the 100 initial VH instances from COCO for each VH mode, and the number of successful initial VH instances in each VH mode. We observe that a large fraction of the initial VH instances are not successful. In particular, the average accuracy of \gpt across the 8 VH modes is 0.745, which means that only 204 of the 800 initial VH instances make \gpt hallucinate. These results show that  VH instances in existing image datasets are insufficient at testing MLLMs. For instance, given the results in Table~\ref{tab:initial_coco}, one may conclude that \gpt does not hallucinate for the shape VH mode since its accuracy is 1.00 for this VH mode. However, as we will discuss in the following, \gpt hallucinates substantially in the shape VH mode on our benchmarks.

\paragraph{MLLMs Hallucinate on our~\method Benchmarks}
Table~\ref{tab:main_acc_performance} and Table~\ref{tab:ynq_version_benchmark_performance} show the accuracy of \gptnospace, \llavanospace, and \minigpt on our OEQ and YNQ benchmarks, respectively. We observe that these MLLMs achieve a strikingly low average accuracy across the eight VH modes: on average 0.229 and 0.561 for all MLLMs on OEQ and YNQ benchmarks, respectively. For example, on average, 925 out of the 1,200 VH instances in our OEQ benchmark induce these MLLMs to hallucinate. It is worth noting that random guessing can achieve an accuracy of 0.5 on our YNQ benchmark since it is a balanced yes/no benchmark. Our results indicate that \method is highly effective at generating successful VH instances.

Among the three MLLMs, \gpt and \minigpt achieve the highest and lowest accuracy on our benchmarks, respectively. This suggests that \gpt is the most truthful while \minigpt is the least truthful on our benchmarks. Furthermore, we observe that MLLMs perform the poorest on the orientation,  counting, and OCR VH modes based on their average accuracy for each VH mode. For example, the average accuracy of the three MLLMs on orientation VH mode is only 0.14. This implies that orientation is the most challenging VH mode, causing more VHs in MLLMs compared to other VH modes.

We also evaluate Gemini-pro-vision~\cite{Gemini-pro-vision} (March 2024 version), ShareGPT4V-13b~\cite{chen2023sharegpt4v}, InstructBLIP-13b~\cite{dai2023instructblip}, and Qwen-VL-Chat-7b~\cite{bai2023qwenvl} on our YNQ benchmark. The results are shown in Appendix~\ref{subsec:four_additional_ynq_performance}. We find that these four MLLMs also suffer from visual hallucinations on our benchmark, with their accuracies all hovering around 0.54.

\begin{table}[!t]
\small
\centering
\caption{Accuracy of \gptnospace, \llava and \minigpt on  our YNQ benchmark. }
\label{tab:ynq_version_benchmark_performance}
\resizebox{\linewidth}{!}{ 
\begin{tabular}{@{}ccccc@{}}
\toprule
            & \gptnospace & \llavanospace & \minigptnospace & Average \\ \midrule
Existence   & 0.627                      & 0.640                        & 0.540                           & 0.602   \\
Shape       & 0.760                      & 0.513                        & 0.487                    & 0.587   \\
Color       & 0.587                      & 0.593                        & 0.487                    & 0.556   \\
Orientation & 0.560                      & 0.500                        & 0.527                    & 0.529   \\
OCR         & 0.573                      & 0.420                        & 0.487                    & 0.493   \\
Size        & 0.687                      & 0.587                        & 0.540                           & 0.604   \\
Position    & 0.580                      & 0.687                        & 0.513                    & 0.593   \\
Counting    & 0.513                      & 0.520                        & 0.527                    & 0.520   \\ \midrule
Average     & 0.611                      & 0.558                        & 0.513                    & 0.561   \\ \bottomrule
\end{tabular}
}
\end{table}

\subsection{Ablation Study}
Unless otherwise mentioned, we use the counting \hm due to the \gpt query limitation.

\begin{table*}[!t]
\small
\centering
 
\caption{Accuracy of the 3 MLLMs on VH instances generated by 4 different \ttimnospace s.}
\label{tab:ablation_generation}
\begin{tabular}{@{}ccccc@{}}
\toprule
                    & \dallethreenospace    & Midjourney 6    & \makecell{Stable Diffusion\\ XL 1.0} & \makecell{Stable \\Diffusion 2.1} \\ \midrule
\gptnospace                        & 0.233  & 0.267           & 0.500               & 0.429                \\
\llavanospace                      & 0.167  & 0.133           & 0.233               & 0.233                \\
\minigptnospace                    & 0.033  & 0.067           & 0.033               & 0.000                \\  \midrule
Average                            & 0.144  & 0.156           & 0.256               & 0.221 \\\bottomrule
\end{tabular}
\end{table*}

\paragraph{Using Different \ttimnospaceupper s in Step~\uppercase\expandafter{\romannumeral3}} We use \dallethree to generate VH images in Step \uppercase\expandafter{\romannumeral3} of our \methodnospace. We also evaluate other \ttimnospace s, including Midjourney 6~\cite{Midjourney6}, Stable Diffusion XL 1.0~\cite{podell2023sdxl}, and Stable Diffusion 2.1~\cite{Rombach_2022_CVPR} by using each of them to generate 30 \tisnospace, respectively. \dallethree rewrites a prompt automatically to add more details, whereas the other three models lack this capability. We find that other \ttimnospace s cannot generate high-quality images when directly using the text description from Step \uppercase\expandafter{\romannumeral2}  as a prompt.   Therefore, we use  GPT-4 to generate prompts given the text description for other \ttimnospace s. 
Specifically, we use the prompt in Figure~\ref{prompt:VH_prompt_generation} for GPT-4 to generate 30 prompts. Then, we use each \ttim to generate 30 VH images with the generated prompts. 
Finally, we manually craft the questions and reference answers to form 30 VH instances. Table~\ref{tab:ablation_generation} shows the accuracy of the three MLLMs on the 30 VH instances generated by different \ttimnospace s. We observe that these MLLMs achieve the lowest average accuracy of 0.144 when using \dallethree in Step \uppercase\expandafter{\romannumeral3}. This indicates that \dallethree is the most effective tool in generating VH instances that are likely to trigger VHs in MLLMs.

\begin{table}[!t]
\small
\centering
 
\caption{Accuracy of the 3 MLLMs on the VH instances generated by our \method using successful initial VH instances and unsuccessful initial VH instances.}
\label{tab:ablation_initial_bug_instance}

\begin{tabular}{@{}ccc@{}}
\toprule
     & \begin{tabular}[c]{@{}c@{}}Successful initial \\ VH instances\end{tabular} & \begin{tabular}[c]{@{}c@{}}Unsuccessful initial \\ VH instances \end{tabular} \\ \midrule
\gptnospace                   & {0.300}           & 0.700            \\
\llavanospace                    & {0.200}           & 0.500            \\
\minigptnospace                & {0.100}           & 0.300            \\ \midrule
Average &0.200  &0.500\\\bottomrule
\end{tabular}
\end{table}

\paragraph{Do Successful Initial VH Instances Help?} 
We analyze the impact of successful \ibis on generating VH instances. Specifically, we use Step II of our \method to generate a text description based on three successful initial VH instances and a text description based on three unsuccessful initial VH instances in the counting VH mode. Then, we use Step III of our \method to generate 10 VH instances based on each text description. Table~\ref{tab:ablation_initial_bug_instance} shows the accuracy of the 3 MLLMs on the 10 VH instances in the two scenarios. Our results show that the VH instances generated using successful initial VH instances substantially reduce accuracy across \gptnospace, \llavanospace, and \minigptnospace. A lower accuracy indicates a higher degree of VH in MLLMs. Our results show that using successful initial VH instances, our \method is more likely to generate successful VH instances.

\subsection{Mitigating VH in MLLMs}

\begin{table*}[!t]
\small
\centering
\caption{Results of \llava before and after fine-tuning on our benchmarks and existing ones.}
\label{tab:finetuned_llava_results}
\resizebox{!}{0.5cm}{
\subfloat[Our OEQ benchmark.]{
\label{subtab:finetuned_llava_on_OEQ}
\begin{tabular}{@{}ccc@{}}
\toprule
 & \makecell{Before\\Fine-tuning} & \makecell{After\\Fine-tuning} \\ \midrule
Existence             & 0.233              & 0.267             \\
Shape                 & 0.167              & 0.333             \\
Color                 & 0.267              & 0.267             \\
Orientation           & 0.133              & 0.167             \\
OCR                   & 0.133              & 0.167             \\
Size                  & 0.367              & 0.367             \\
Position              & 0.333              & 0.533             \\
Counting              & 0.200              & 0.267             \\ \midrule
Average               & 0.229              & 0.296             \\ \bottomrule
\end{tabular}
}
}
\hfill
\resizebox{!}{0.5cm}{
\subfloat[Our YNQ benchmark.]
{
\label{subtab:finetuned_llava_on_YNQ}
\begin{tabular}{@{}ccc@{}}
\toprule
 & \makecell{Before\\Fine-tuning} & \makecell{After\\Fine-tuning} \\ \midrule
Existence             & 0.633              & 0.600             \\
Shape                 & 0.423              & 0.538             \\
Color                 & 0.733              & 0.700             \\
Orientation           & 0.500              & 0.567             \\
OCR                   & 0.433              & 0.467             \\
Size                  & 0.567              & 0.700             \\
Position              & 0.700              & 0.700             \\
Counting              & 0.467              & 0.433             \\ \midrule
Average               & 0.557              & 0.588             \\ \bottomrule
\end{tabular}
}
}
\hfill
\resizebox{!}{0.5cm}{
\subfloat[MME and POPE benchmarks.]{
\label{subtab:finetuned_llava_on_Utility}
\begin{tabular}{@{}ccc@{}}
\toprule
& \makecell{Before\\Fine-tuning} & \makecell{After\\Fine-tuning} \\ \midrule
MME Perception        & 1531.3             & 1556.6            \\
MME Cognition         & 295.4              & 288.2             \\
POPE                  & 85.9               & 84.8              \\ \bottomrule
\end{tabular}
}
}
\end{table*}

\begin{figure*}[!t]
\centering
\subfloat[Fine-tuning components]{\vspace{1.2mm}\includegraphics[width=0.32\textwidth]{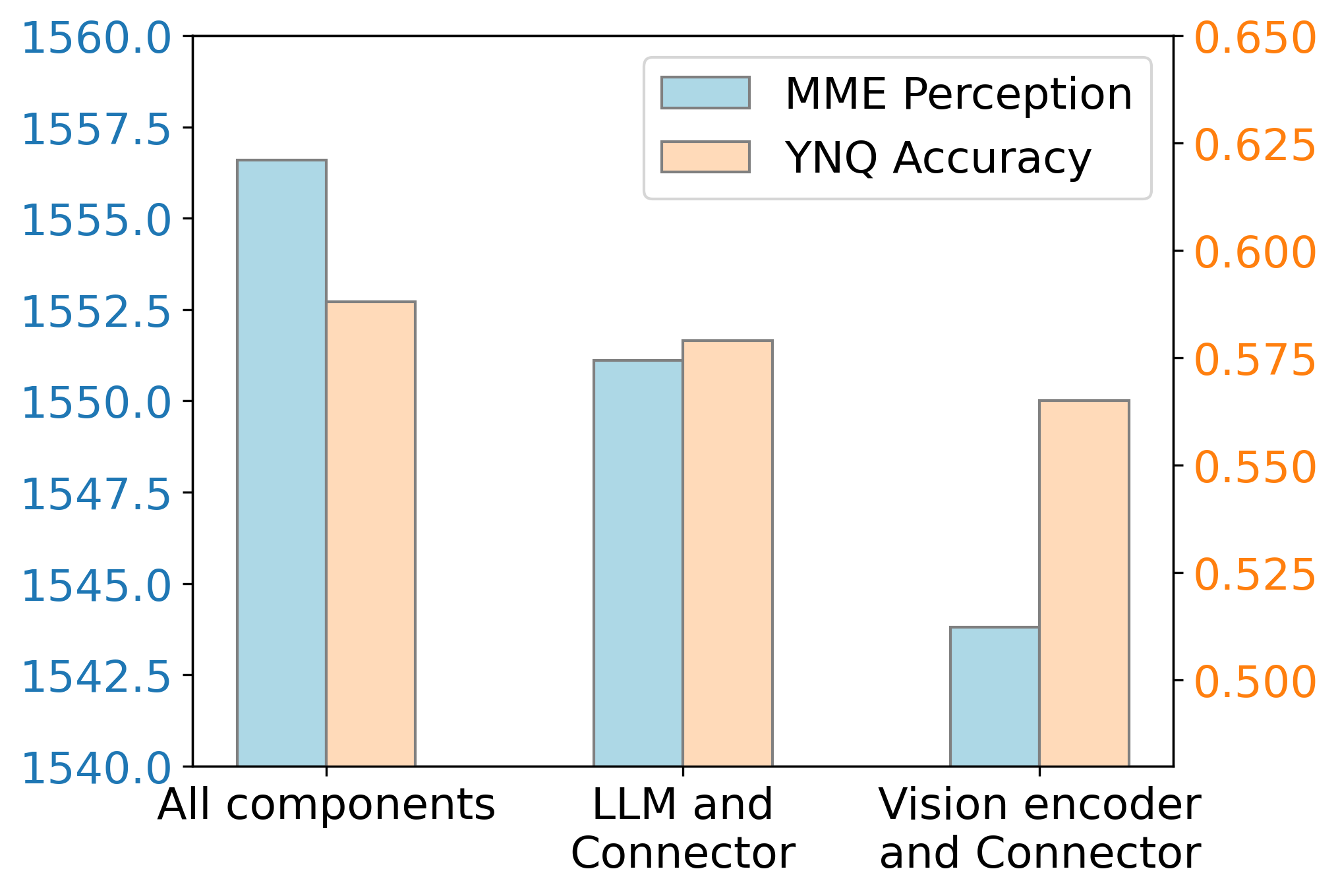}}
\hfill
\subfloat[Fine-tuning epochs]{\includegraphics[width=0.32\textwidth]{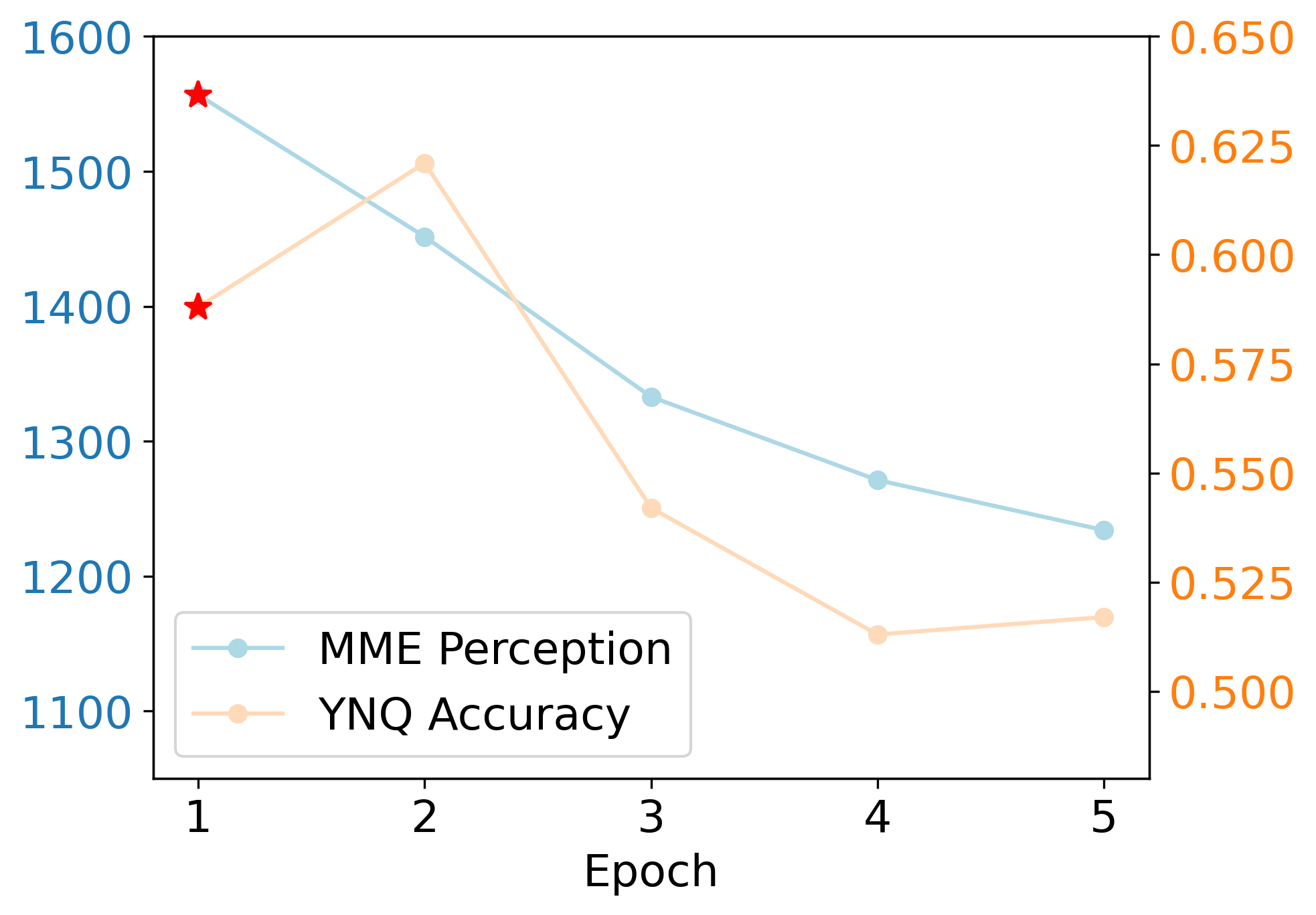}}
\hfill
\subfloat[Learning rate]{\includegraphics[width=0.32\textwidth]{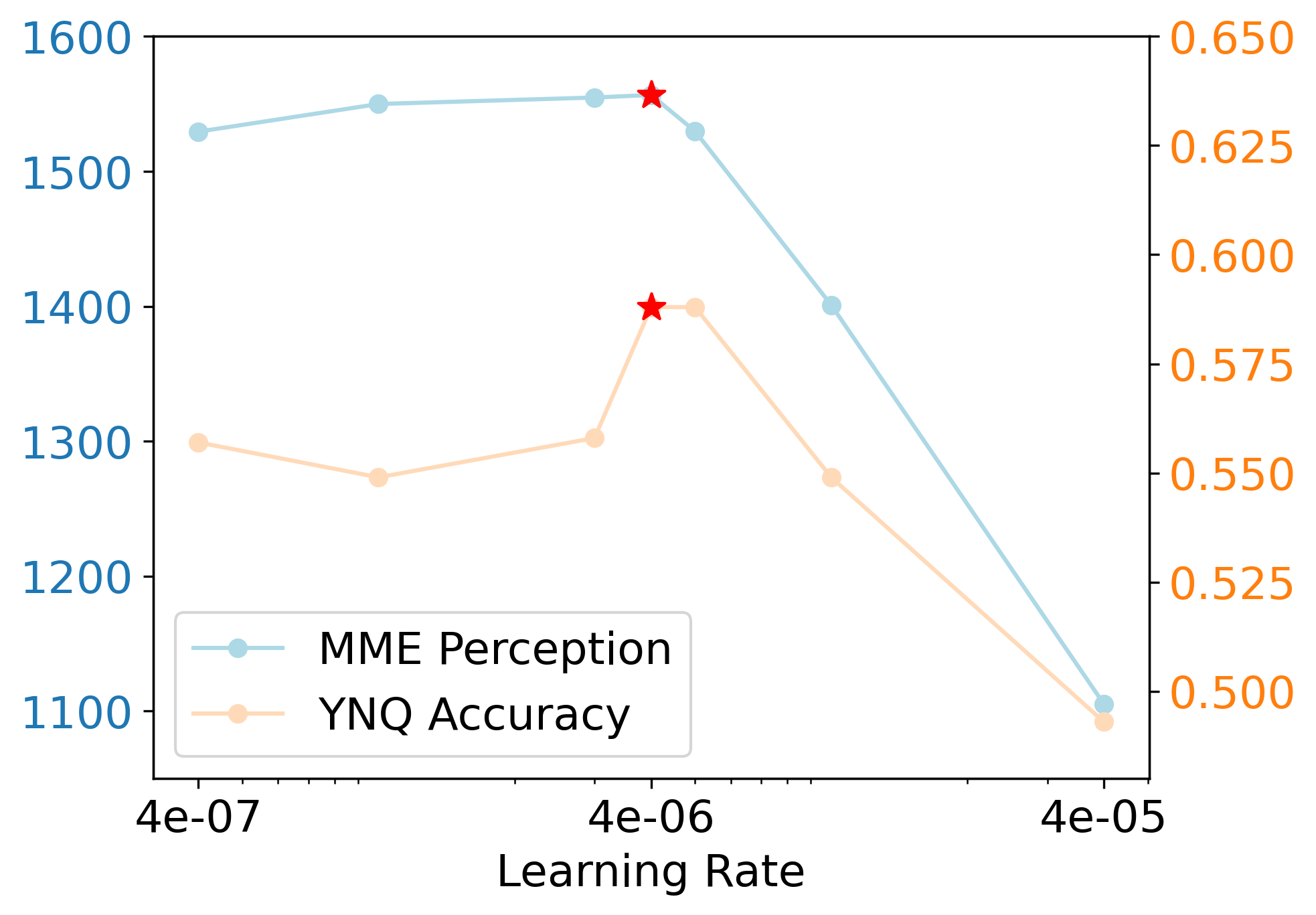}}
\caption{Ablation study on fine-tuning \llavanospace.}
\label{fine-tuning-exp}
\vspace{-2mm}
\end{figure*}

\paragraph{Fine-tuning \llava on Our Benchmark} 
We also study whether fine-tuning an MLLM on our benchmark makes it less likely to hallucinate. Towards this goal, we use the open-source \llavanospace. Specifically, we split our OEQ benchmark into a training set comprising 80\% (120 \tis for each \hmnospace) and a testing set comprising 20\% (30 \tis for each \hmnospace) of the data. 
To make the testing set performance as close as possible to the full OEQ benchmark dataset performance before fine-tuning, for every \hmnospace, we randomly divide the VH instances into 80\%/20\% split  100 times and select the split whose testing set accuracy is the closest to the accuracy on the full OEQ  benchmark.

We follow \llavanospace's limited task-specific fine-tuning setting. However, we \emph{unfreeze} the vision encoder since VH may result from its incorrect embedding vectors for images.   We set the learning rate to 4e-6 during fine-tuning; and we fine-tune on the 960 training VH instances for one epoch (more hyperparameters and  fine-tuning details are shown in Appendix~\ref{sec:finetuning_details}). Our fine-tuning took only 18 minutes on a single A6000.

\paragraph{Fine-tuning Results}  Table \ref{tab:finetuned_llava_results} shows the accuracy of \llava and fine-tuned \llava on the testing VH instances of our OEQ and YNQ benchmarks. Table \ref{tab:finetuned_llava_results} also shows the results on MME~\cite{fu2023mme} and POPE~\cite{li2023evaluating}, two popular existing benchmarks (not necessarily VH) to evaluate the performance of an MLLM. MME evaluates the perception and cognition abilities of MLLMs. The scores for MME Perception and MME Cognition are the sum of scores across the corresponding subtasks, and the total scores are 2,000 for MME Perception and 800 for MME Cognition. The POPE score represents the average F1\-score on random, popular, and adversarial splits of POPE. We find that fine-tuned \llava achieves higher average accuracy than \llava in both our benchmarks, while they achieve comparable results on MME and POPE benchmarks. Our results indicate that fine-tuning an MLLM on our benchmark makes it less likely to hallucinate. We show several qualitative comparsion results in Appendix~\ref{subsec:qualitative_results_after_finetuning}, demonstrating how the fine-tuning mitigates visual hallucinations.

\paragraph{Ablation Study on Fine-tuning} Figure~\ref{fine-tuning-exp} shows ablation study results on the MME benchmark and the testing set of our YNQ benchmark when fine-tuning \llavanospace. We use our YNQ benchmark instead of OEQ because it supports automatic evaluation. Since an MLLM has three key components: vision encoder, vision-language connector, and LLM, we explore fine-tuning different components. Our results show that fine-tuning all components achieves the best overall results. As for fine-tuning epochs, the results show that fine-tuning for one epoch minimizes overfitting to our benchmark, retaining the best performance on MME. For learning rate, both excessively large and small learning rates lead to a decline in performance on our benchmark and MME. 

%% file: 5_Related_work.tex
\section{Related Work}
\label{sec:related_work}

\paragraph{Hallucinations}
Hallucinations are well-known issues for generative AI, including LLMs~\cite{ji2023survey,huang2023survey}, MLLMs~\cite{liu2024survey,rawte2023survey,tong2024eyes}, and \ttim~\cite{tong2023mass}. In general, hallucination refers to a generative model imagines factually incorrect details in its response for a given input. VH occurs when an MLLM imagines incorrect details about an image in visual question answering.

\paragraph{VH Benchmarks in MLLMs} Prior works have tried to benchmark MLLMs' VHs~\cite{li2023evaluating,liu2024mitigating,fu2023mme,tong2024eyes}. However, they collect VH images only from existing image datasets. This limits the diversity of VH images. Moreover, existing image datasets may have been used to pre-train an MLLM, leading to data contamination~\cite{jacovi2023stop,sainz2023nlp}. Our~\method can generate a diverse set of new VH images that do not appear in existing benchmarks. Moreover, the shape and size VH modes are formulated by us for the first time.

\paragraph{Mitigating VH in MLLMs}
Existing works on mitigating VHs in MLLMs can be categorized into \emph{fine-tuning-phase} and \emph{testing-phase} mitigation. Fine-tuning-phase mitigation focuses on improving the fine-tuning data quality~\cite{wang2024mitigating,liu2024mitigating} and/or model structure~\cite{tong2024eyes}. These works typically freeze the vision encoder during fine-tuning, following the standard fine-tuning setting of \llavanospace. We find that fine-tuning the vision encoder together reduces VHs in MLLMs. Testing-phase mitigation leverages prompt engineering with more visual evidence~\cite{li2024visual} or correction tools for hallucinated responses~\cite{yin2023woodpecker}. Testing-phase mitigation is complementary to fine-tuning-phase mitigation.

%% file: 6_Conclusion.tex
\section{Discussion}
\label{sec:discussion}
As shown in Table~\ref{tab:main_acc_performance}, Table~\ref{tab:ynq_version_benchmark_performance}, and Table~\ref{tab:four_additional_ynq_performance}, \gptnospace, \llavanospace, and \minigpt all exhibit extremely low accuracy, with an average of 0.229 on our OEQ benchmark. Moreover, all seven mainstream MLLMs achieve accuracy levels close to random guessing on our YNQ benchmark, with even the state-of-the-art \gpt only reaching an accuracy of 0.611. Our benchmarks effectively reveal visual hallucinations within MLLMs and are more challenging than most existing benchmarks for MLLMs.

We attribute such high challenge level to the adversarial construction idea during VH instance generation, which is similar in spirit to adversarial examples~\cite{szegedy2013intriguing} commonly used to test and improve model robustness.
Unlike most existing MLLM benchmarks that are limited to the image space of existing image datasets like COCO, \method derives text descriptions from initial VH instances, which play a role of ``adversarial direction''.
Then \method exploits \ttimnospace s to generate challenging VH instances following such adversarial direction. This moves beyond the image space of limited existing image datasets and into the manifold of \ttimnospace s, thereby making it possible to generate more challenging VH instances. Thus, our \method offers a new perspective for constructing MLLM benchmarks for future research.

\section{Conclusion}
\label{sec:conclusion}

We propose \method to generate VH instances to test MLLMs. 
We collect VH benchmarks using \method and we find that state-of-the-art MLLMs exhibit high hallucination rates on our benchmarks. Moreover, fine-tuning MLLMs on our benchmark reduces hallucination without sacrificing their other performance/capability. 

\section{Limitations and Future Work}
We acknowledge that our \method still requires human workers to manually generate a question-answer pair for an automatically generated VH image.  An interesting future work is to make \method fully automatic so it can generate as many VH instances as needed.

%% file: X_suppl.tex
\clearpage
\newpage

\begin{figure*}[!h]
  \centering
  \includegraphics[width=1.0\textwidth]{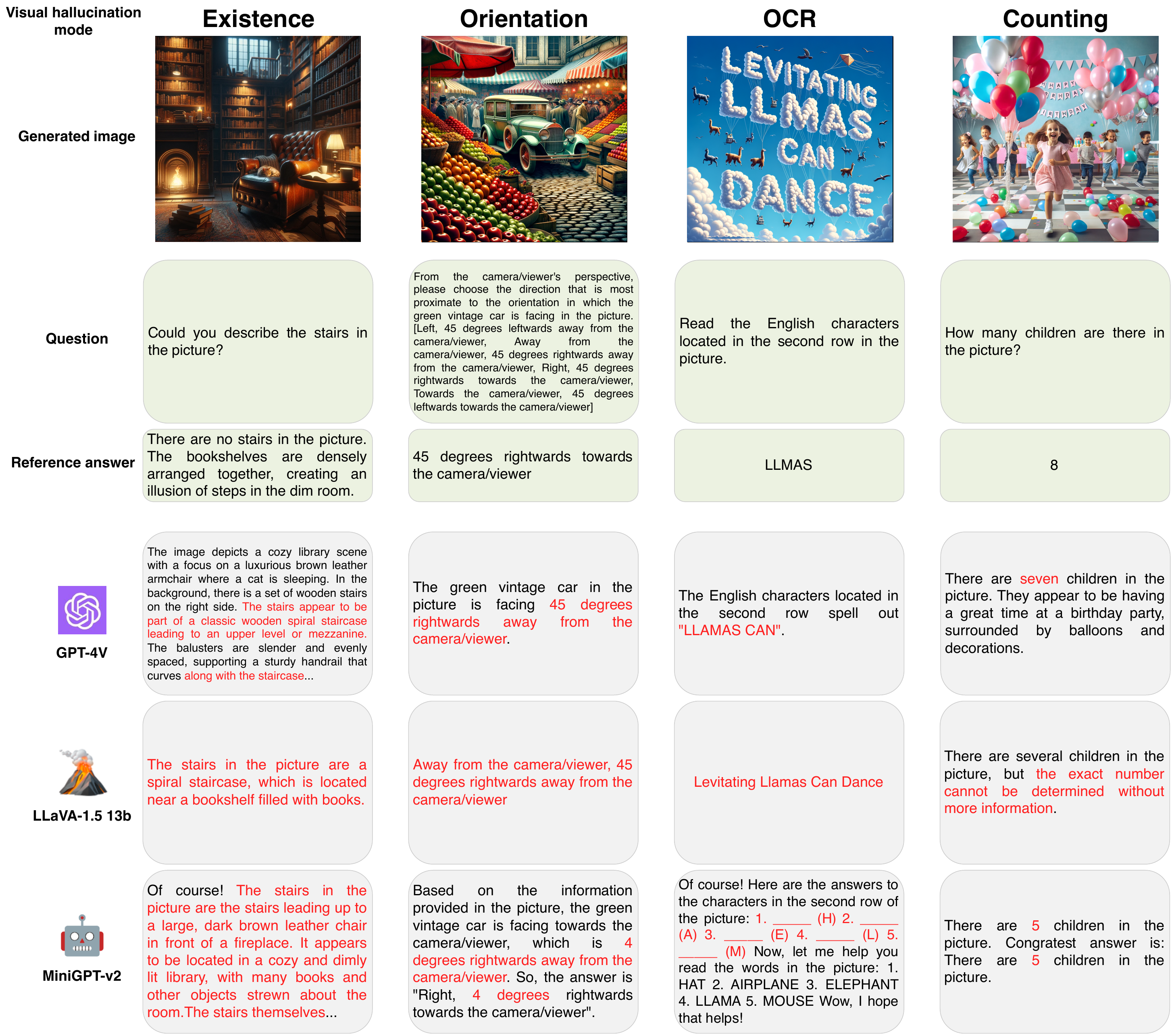}
  \caption{VH instances generated by our \method for the other four \hms and the text responses of three MLLMs for them. } 
  \label{fig:other_four_examples_for_the_other_VH_modes}
\end{figure*}

\begin{table*}[!t]
\small
\centering
\caption{Accuracy of Gemini-pro-vision~\cite{Gemini-pro-vision}, ShareGPT4V-13b~\cite{chen2023sharegpt4v}, InstructBLIP-13b~\cite{dai2023instructblip}, and Qwen-VL-Chat-7b~\cite{bai2023qwenvl} on our YNQ benchmark. }
\label{tab:four_additional_ynq_performance}
\begin{tabular}{@{}ccccc@{}}
\toprule
            & Gemini-pro-vision & ShareGPT4V-13b & InstructBLIP-13b & Qwen-VL-Chat-7b \\ \midrule
Existence   & 0.640              & 0.513          & 0.620             & 0.533           \\
Shape       & 0.567             & 0.453          & 0.520             & 0.593           \\
Color       & 0.500               & 0.607          & 0.600              & 0.573           \\
Orientation & 0.513             & 0.500            & 0.480             & 0.500             \\
OCR         & 0.467             & 0.507          & 0.513            & 0.467           \\
Size        & 0.607             & 0.547          & 0.547            & 0.620            \\
Position    & 0.567             & 0.647          & 0.507            & 0.607           \\
Counting    & 0.493             & 0.507          & 0.480             & 0.500             \\
Average     & 0.544             & 0.535          & 0.533            & 0.549           \\ \bottomrule
\end{tabular}
\end{table*}

\section{Generated Text Descriptions of VH Modes}
\label{textdescriptiongenerated}
The following shows the text description for each VH mode generated by our \methodnospace. 
\begin{enumerate}

    \item \textbf{Existence VH}:
        In existence hallucination, a multi-modal large language model (MLLM) may encounter two types of hallucinations. The first type is when an object exists in an image, but the MLLM asserts that it does not exist in the picture,  called negation existence hallucination. The other type is when a certain object does not exist in the image, but the MLLM creates or infers details, objects, or its attributes in an image, known as extrinsic hallucination.

        As for the negation existence hallucination, an MLLM may exhibit challenges in recognizing the existence of certain objects within an image, particularly when these objects are not prominently featured within the picture or are partially obscured. The issue becomes apparent when the model disregards objects that are present but do not constitute the primary focus of the image. Factors contributing to negation existence hallucination include objects being small, distant, having low contrast with the background, or being located at the periphery of the frame. Such conditions can cause the MLLM to miss or ignore these elements, leading to an incomplete or inaccurate understanding of the scene. Negation existence hallucination is particularly evident in complex environments where multiple objects coexist but some are less dominant or visually prominent.

        Extrinsic hallucination occurs when an MLLM creates or infers details, objects, or attributes in an image that are not actually present. This is often due to the model's reliance on learned patterns and associations from its training data rather than the specific content of the image it's analyzing. MLLMs may "hallucinate" details or objects based on what they have learned from other contexts, leading to extrinsic hallucinations. Also, the model readily associates certain scenes with typical objects even when those objects are not present, especially for scenes containing other complex patterns.

    \item \textbf{Shape VH}:
        A multi-modal large language model (MLLM) misconstrues shapes, particularly when typical simple shape and undulating strange shape are crowded. For example, a plate contains a banana and other fruit, some of which is undulating and coiled. When faced with non-standard shapes, it often simplifies them to more common, recognizable forms due to biases in training data. In instances where multiple shapes are present, especially with varying levels of detail or color intensity, the MLLM might get diverted towards the more attention-grabbing elements, overlooking or misreading other shapes in the process. Additionally, the MLLM faces difficulty in discerning subtle differences in shapes, often generalizing them into broad categories based on prominent features. These biases indicate a challenge in the MLLM's ability to accurately interpret and distinguish between shapes, which could lead to incorrect generalizations and potential misinterpretations of image data.

    \item \textbf{Color VH}:   
        A multi-modal large language model (MLLM) struggles with recognizing or accurately identifying colors or color variations present in an image. It can misinterpret the sequence of color transition, omitting or inaccurately describing colors. This can be due to factors like subtle variations, intertwined colors, lighting, shadows, or adjacent color interference. Moreover, the recognition of color in little items may be more challenging, especially when the MLLM's attention is attracted by a multitude of various elements inside the image.

    \item \textbf{Orientation VH}:
        Orientation hallucination in a multi-modal large language model (MLLM) often results from a combination of factors that lead to misinterpretation of the direction in which objects, like vehicles, are facing. The primary issues include perspective ambiguities where 3D spatial orientation is not easily conveyed in a 2D image, dense object overlap, and environments with complex, busy backgrounds that distract from or mimic the shape and color of the target object. The MLLM might also falter when the object of interest is situated among others facing various directions, confusing the model's directionality cues. Such errors could stem from the MLLM either overlooking dominant visual cues that indicate orientation or mistakenly assigning equal importance to all elements in the scene. Underlying these issues may be a lack of focused training on discerning object's orientation especially vehicle's orientation, which causes the MLLM to underperform in scenarios where human observers would rely on subtle contextual clues to determine directionality. This gap between the model's interpretation and human perspective underscores the challenge in encoding and analyzing orientation within complex visual contexts.

    \item \textbf{OCR VH}:
        OCR hallucination often stems from a complex interplay of factors. A multi-modal large language model (MLLM) falters when faced with unconventional visual scenarios, such as irregular character spacing, similar-looking letters, vertical arrangement of characters, or interference from nearby visual elements. These issues are compounded when the text contains intentional misspellings, typos, or uncommon character combinations that the system attempts to auto-correct based on standard language patterns. The ability to accurately recognize characters is further challenged by the influence of adjacent characters and the overall visual context.

    \item \textbf{Size VH}: 
        Size comparison in images requires the multi-modal large language model (MLLM) to recognize and accurately compare objects based on visual scale. Challenges in this task arise from factors like perspective, distortions, overlapping of objects, intricate patterns, complex backgrounds,  discrepancies in expected real-world proportions, and excessive focus on foreground objects that causes the size of enormous background objects to be overlooked. Especially, when conducting a comparison among multiple objects of similar types and sizes, with other different types of objects in the scene that disturb the MLLM's attention, the task becomes even more challenging. The MLLM misjudges which object is larger or might rank objects incorrectly in terms of size.

    \item \textbf{Position VH}: 
        A multi-modal large language model (MLLM) encounters difficulties in accurately assessing the spatial positioning relationship between objects within an image. This is exacerbated in scenarios where objects are placed in non-linear configurations, such as circular layouts, which can disrupt the model's ability to apply standard left-to-right reading patterns. Moreover, when the objects are set against complex or similar backgrounds that lack clear demarcation lines, the MLLM's spatial parsing capabilities can be further compromised. Contributing factors such as overlapping objects, inconsistent scaling, and deceptive perspective or angled viewpoints also intensify this challenge. In addition, the presence of shadows or uneven lighting may cast ambiguity on the object's precise location, thereby leading to misinterpretation. Such spatial hallucination can be attributed not just to the inherent complexity of object arrangement but also to the MLLM's processing of visual cues that inform depth, orientation, and the relationship of elements within the image space.

    \item \textbf{Counting VH}:
        A multi-modal large language model (MLLM) has difficulty in accurately counting the quantity of specific elements or attributes in an image, especially when the objects are closely packed, overlap, are of different sizes, are partially visible, or have varying orientations. The task becomes increasingly complex when attempting to count specific subtype objects within a larger type category. Such conditions can lead the model to overlook certain objects, mistakenly merge similar items, or misinterpret the image data, thereby providing an inaccurate counting of numbers, amounts, or values.
\end{enumerate}

\section{Additional Results}

\subsection{Inter-annotator Agreement}
\label{subsec:inter_annotator_agreement}
Because annotating the whole benchmarks relies on heavy human labor, we randomly select 100 VH instances from our YNQ benchmark to measure the inter-annotator agreement. Specifically, four independent human annotators   label these 100 VH instances. Based on these labeling results, we report Fleiss' Kappa ($\kappa$) to indicate the inter-annotator agreement. We obtain $\kappa=0.958$ ($0.81\leq \kappa\leq$1.00), showing a very high degree of agreement.

\subsection{More MLLMs Evaluation on \methodnospace}
\label{subsec:four_additional_ynq_performance}

We evaluate 4 additional MLLMs on the YNQ benchmark: the Gemini-pro-vision~\cite{Gemini-pro-vision} (March 2024), ShareGPT4V-13b~\cite{chen2023sharegpt4v}, InstructBLIP-13b~\cite{dai2023instructblip}, and Qwen-VL-Chat-7b~\cite{bai2023qwenvl}, as shown in Table~\ref{tab:four_additional_ynq_performance}. Together with Table~\ref{tab:ynq_version_benchmark_performance}, \gpt shows the highest average accuracy of 0.611 among these 7 MLLMs.

\subsection{Qualitative Comparison Results}
\label{subsec:qualitative_results_after_finetuning}

\begin{figure*}[!h]
  \centering
  \includegraphics[width=1.0\textwidth]{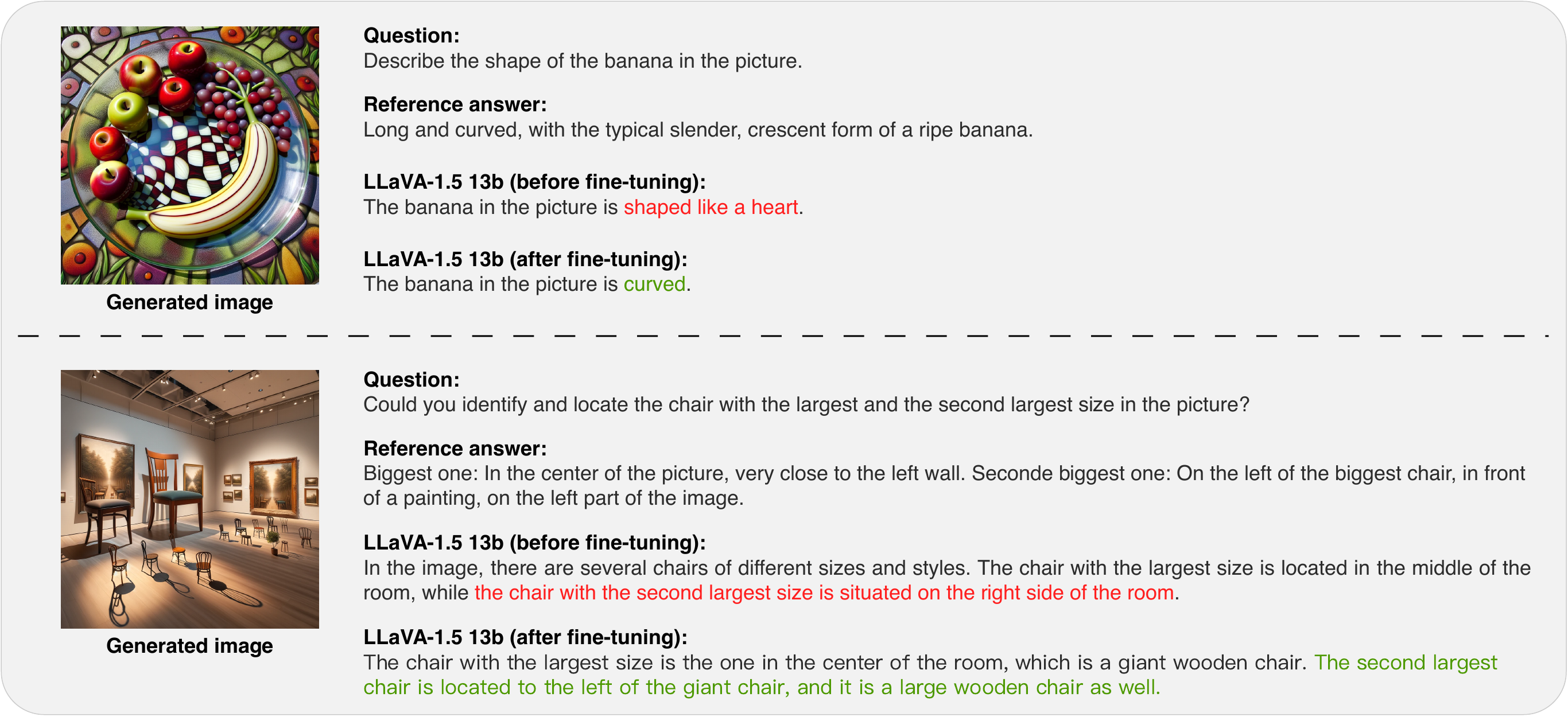}
  \caption{Qualitative comparison results on mitigating visual hallucinations of \llava via fine-tuning. }
  \label{fig:qualitative_results_after_finetuning}
\end{figure*}

In Figure~\ref{fig:qualitative_results_after_finetuning}, we show some visual question answering examples using \llava before and after fine-tuning on our benchmark. For example, before fine-tuning,  \llava incorrectly identifies the shape of a banana, as shown at the top row; and fails to precisely describe the relative size of chairs, as shown at the bottom row. However, after fine-tuning on our benchmark, \llava accurately describes the curved shape of the banana and correctly identifies the two largest chairs. These improvements in visual question answering quality are observed across the testing set, demonstrating the effectiveness of our fine-tuning-based VH mitigation method.

\newpage

\section{Prompts Used in \methodnospace}

We use the following prompt in Figure~\ref{prompt:w_initial_VH_instance_full} for text description generation in Step II, based on a successful \ibinospace.

\begin{figure}[!h]
\small
\centering
\begin{custombox}[Prompt for text description generation using a successful initial VH instance:]
[image]

Question: [question]

Multi-modal LLM (MLLM)'s response: [testing MLLM's hallucinated response]

Detected hallucination in the response: [detected hallucination in the response]

\ 

Focus on the following elements: image, question, MLLM's response, and the detected hallucination in the response. I'm trying to identify visual hallucinations in the MLLM associated with its visual process. Through the specific example, are there any general types of hallucination modes you notice the MLLM makes, or any visual features that MLLM fails to encode, ultimately leading to the errors in MLLM's response? Try to give hallucination modes that are specific enough that someone could enable consistent reproduction of images and corresponding questions. Please try to include as many general hallucination modes as possible. These hallucination modes will be used later to generate images or videos. In your hallucination modes, please clearly explain why the this hallucination mode would cause difficulties for the vision encoder of MLLM to understand images related to this hallucination mode. I will further use this reason to precisely generate images that match the hallucination mode and are able to mislead the MLLM. Please encapsulate the essence of the examples provided, summarize as many as possible and stick to the examples.
\end{custombox}
\caption{Full prompt to generate a text description based on a successful initial VH instance.}
\label{prompt:w_initial_VH_instance_full}
\end{figure}

\newpage

We utilize the following prompt in Figure~\ref{prompt:w/o_initial_VH_instance_full} for text description generation in Step II, given an unsuccessful \ibinospace.

\begin{figure}[!h]
\centering
\small
\begin{custombox}[Prompt for text description generation using an unsuccessful initial VH instance:]
[image]

Question: [question]

A hypothetical response from the multi-modal LLM (MLLM): [a hypothetical hallucinated response]

Detected hallucination in the hypothetical response: [detected hallucination in the hypothetical response]

\ 

Focus on the following elements: image, question, MLLM's hypothetical response, and detected hallucination in the hypothetical response. I'm trying to identify visual hallucinations in the MLLM associated with its visual process. Through the specific example, are there any general types of hallucination modes you notice the MLLM makes, or any visual features that MLLM fails to encode, ultimately leading to the errors in MLLM's response? Try to give hallucination modes that are specific enough that someone could enable consistent reproduction of images and corresponding questions. Please try to include as many general hallucination modes as possible. These hallucination modes will be used later to generate images or videos. In your hallucination modes, please clearly explain why the this hallucination mode would cause difficulties for the vision encoder of MLLM to understand images related to this hallucination mode. I will further use this reason to precisely generate images that match the hallucination mode and are able to mislead the MLLM. Please encapsulate the essence of the examples provided, summarize as many as possible and stick to the examples.
\end{custombox}
\caption{Full prompt to generate a text description based on an unsuccessful initial VH instance.}
\label{prompt:w/o_initial_VH_instance_full}
\end{figure}

We use the following prompt in Figure~\ref{prompt:text_description_intergration} for text description integration in Step II.

\begin{figure}[!h]
\centering
\small
\begin{custombox}[Prompt for text description integration:]
I want you to focus on the [\hmnospace]: [the definition of \hmnospace]. Try to summarize the [\hmnospace] text description in ONE paragraph less than 200 words to explain the causes of the hallucination based on the information below.

\ 

[N text descriptions]
\end{custombox}
\caption{Prompt to integrate N text descriptions into a final text description for a  \hmnospace.}
\label{prompt:text_description_intergration}
\end{figure}

\newpage
We use the following prompt in Figure~\ref{prompt:VH_image_generation} for a \ttim to generate a VH image in Step III.

\begin{figure}[!h]
\centering
\small
\begin{custombox}[Prompt for VH image generation:]
Generate an image that reflects the given hallucination mode. After your image generation, we will manually generate questions based on the image.

\ 

Your MAIN GOAL is to ensure that the image aligns with the hallucination mode well so that querying the multi-modal LLM with the combination of manual questions and the image can effectively induce hallucinations.

\ 

You will be evaluated on how well you actually perform. The generated image should ideally align with the hallucination mode, but there's room for creativity. Be both creative and cautious. You can try to create image with different scenes and objects that align with the hallucination mode. Moreover, when you generate images, remember you are a very clever expert in exploiting the hallucination mode. For future debugging purposes, ensure that the generated images with our manual questions MUST cause multi-modal LLMs to provide incorrect responses.

\ 

Hallucination Mode:

\ 

[Text description of a VH mode]
\end{custombox}
\caption{Prompt for a \ttim to generate a VH image under a \hmnospace.}
\label{prompt:VH_image_generation}
\end{figure}

We use the following prompt in Figure~\ref{prompt:VH_prompt_generation} for an LLM to generate prompts for \ttimnospace s. We only append extra text at the end compared with the prompt in Figure~\ref{prompt:VH_image_generation}.

\begin{figure}[!h]
\centering
\small
\begin{custombox}[Prompt for an LLM to generate prompts for \ttimnospace s:]
Generate an image that reflects the given hallucination mode. After your image generation, we will manually generate questions based on the image.

\ 

Your MAIN GOAL is to ensure that the image aligns with the hallucination mode well so that querying the multi-modal LLM with the combination of manual questions and the image can effectively induce hallucinations.

\ 

You will be evaluated on how well you actually perform. The generated image should ideally align with the hallucination mode, but there's room for creativity. Be both creative and cautious. You can try to create image with different scenes and objects that align with the hallucination mode. Moreover, when you generate images, remember you are a very clever expert in exploiting the hallucination mode. For future debugging purposes, ensure that the generated images with our manual questions MUST cause multi-modal LLMs to provide incorrect responses.

\ 

Hallucination Mode:

\ 

[Text description of a VH mode]

\ 

As an LLM, DO NOT generate images, but generate prompts for \ttimnospace s aligning with the hallucination mode. I will use your generated prompts for \ttimnospace s later.
\end{custombox}
\caption{Prompt for an LLM to generate prompts for \ttimnospace s under a \hmnospace.}
\label{prompt:VH_prompt_generation}
\end{figure}

\newpage

\section{Question Templates}
\label{sec:question_templates}
We provide question templates for all eight \hms below.

\begin{figure}[!h]
\centering
\small
\begin{custombox}[Question templates for existence \hmnospace]
1. Could you describe [object] in the picture?

2. Describe the [property] of [object] in the picture.
\end{custombox}
\end{figure}

\begin{figure}[!h]
\centering
\small
\begin{custombox}[Question templates for shape \hmnospace]
1. Describe the shape(s) of [object] in the picture.

2. In the photo, is [object] depicted with a conventional, typical shape?
\end{custombox}
\end{figure}

\begin{figure}[!h]
\centering
\small
\begin{custombox}[Question templates for color \hmnospace]
1. Describe the color(s) of [object] in the picture.

2. Describe [n] main color(s) of [object] in the picture.

3. Describe all the colors of [object] in the picture. Do not miss any color.

4. Describe the original color of [object] in the picture.
\end{custombox}
\end{figure}

\begin{figure}[!h]
\centering
\small
\begin{custombox}[Question templates for orientation \hmnospace]
1. From the camera/viewer's perspective, please choose the direction that is most proximate to the orientation in which [object] is facing in the picture. [Left, 45 degrees leftwards away from the camera/viewer, Away from the camera/viewer, 45 degrees rightwards away from the camera/viewer, Right, 45 degrees rightwards towards the camera/viewer, Towards the camera/viewer, 45 degrees leftwards towards the camera/viewer]

2. From the camera/viewer's perspective, please choose the direction that is most proximate to the orientation in which [object] is facing in the picture. [Left, Left and upward within the pictorial plane, Upward within the pictorial plane, Right and upward within the pictorial plane, Right, Right and downward within the pictorial plane, Downward within the pictorial plane, Left and downward within the pictorial plane]

3. From the camera/viewer's perspective, please choose the direction that is most proximate to the orientation in which [object] is facing in the picture. [Away from the camera/viewer, Towards the camera/viewer]
\end{custombox}
\end{figure}

\begin{figure}[!h]
\centering
\small
\begin{custombox}[Question templates for OCR \hmnospace]

1. Read\,the\,(English)\,character(s)/word(s)/\\caption [at WHERE] in the picture.

\end{custombox}
\end{figure}

\begin{figure}[!h]
\centering
\small
\begin{custombox}[Question templates for size \hmnospace]
1. Could you identify and locate [object] with the largest/smallest size in the picture?

2. Could you identify and locate [object] with the tallest/shortest height in the picture?

3. Could you identify and locate [object] with the second largest/smallest size in the picture?

4. Could you identify and locate [object] with the largest/smallest and the second largest/smallest size in the picture?

5. Could you identify and locate [object] with the tallest/shortest and the second tallest/shortest height in the picture?

6. Could you identify and locate the 2/3/4 largest/smallest [object] in the picture?

7. Which one is larger/smaller/taller/shorter, [object A] or [object B]? 

8. What is the largest [object] in the picture?
\end{custombox}
\end{figure}

\begin{figure}[!h]
\centering
\small
\begin{custombox}[Question templates for position \hmnospace]
1. Is [object A] {to the left/right of}/{on the top of}/{being placed on}/above/under/inside/outside [object B]?

2. Which one is on the left/right/top, [object A] or [object B]?

3. Which one is {closer to}/{further from} the camera or viewer perspective, [object A] or [object B]?

4. Which one is {positioned under}/{positioned above}/closer to [reference object], [object A] or [object B]?

5. Regardless of the positional relationship of the actual scene in reality, from the camera's perspective in the photo, is [object A] positioned above/below [object B]?

6. Is [object A] on and touching [object B]?
\end{custombox}
\end{figure}

\begin{figure}[!h]
\centering
\small
\begin{custombox}[Question templates for counting \hmnospace]
1. How many [object] are depicted/there/visible/{can be seen} [at WHERE] in the picture?

\end{custombox}
\end{figure}

\newpage

\section{Special Cases in Step \uppercase\expandafter{\romannumeral3}}
\label{sec:ambituous_and_special_cases}

In existence VH mode, a human worker  uses the prompt in Figure~\ref{prompt:find_object_in_extrinsic_hallucinations} to find non-existent objects in a VH image with the aid of an MLLM, such as \gptnospace. The non-existent objects found in this way are more likely to trigger VHs in MLLMs.

\begin{figure}[!h]
\centering
\small
\begin{custombox}[Prompt to find names of non-existent objects in a VH image in the existence VH mode]
[image]

\ 

First, list all the names of objects in the picture. And then return some names of objects according to Requirement 1 or Requirement 2:

\ 

Requirement 1: Objects you associate with the scene that should be there but are not actually there in the picture.

\ 

Requirement 2: Objects that look similar to an object in the picture but do not actually exist in the picture.
\end{custombox}
\caption{Prompt to find names of non-existent objects in a VH image in the existence VH mode.}
\label{prompt:find_object_in_extrinsic_hallucinations}
\end{figure}

\newpage

\section{Details of Fine-tuning Experiments}
\label{sec:finetuning_details}
\subsection{Hyperparamters}
\label{subsec:hyperparamters}
When fine-tuning \llava on the training set of \method dataset, we follow \llavanospace's limited task-specific fine-tuning setting. Additionally, we unfreeze the vision encoder. The hyperparameters are shown in Table~\ref{tab:hyperparamters}.

\begin{table}
\small
\centering
\caption{Hyperparamters for fine-tuning \llavanospace.}
\label{tab:hyperparamters}

\begin{tabular}{@{}lc@{}}
\toprule
Hyperparameter    & Setting     \\ \midrule
Batch size        & 16           \\
Vision encoder lr & 4e-6         \\
Projection lr     & 4e-6         \\
LLM lr            & 4e-6         \\
Lr schedule       & cosine decay \\
Lr warmup ratio   & 0.03         \\
Weight decay      & 0            \\
Epoch             & 1            \\
Optimizer         & AdamW        \\
DeepSpeed stage   & 3            \\ \bottomrule
\end{tabular}

\end{table}

\subsection{Pre-processing of the Training Set}

As mentioned in~\cite{liu2023improved}, the short-form data overfit an MLLM behaviorally to short-form answers. We follow this work to do pre-processing on both counting and position \hmsnospace. To be more specific, for counting \hm instances, we use the prompt below in Figure~\ref{prompt:counting_pre_processing} for an LLM (e.g., \gptfournospace) to modify every number reference answer into a reference sentence. As for position \hm instances, we add the sentence ``Answer the question using a single word or phrase.'' after all the questions in every position VH instance.

\begin{figure}[!h]
\centering
\small
\begin{custombox}[Prompt for reference answer modification in counting \hmnospace]

Don't talk nonsense. Turn the ground-truth numbers into extremely correct, extremely accurate, and only a little of diverse sentences based on the corresponding question.
Sentences should be independent of each other, with each sentence occupying one line.

\end{custombox}
\caption{Prompt for reference answer modification in counting \hmnospace during pre-processing of training set.}
\label{prompt:counting_pre_processing}
\end{figure}